\title{ARXiV--Automating Reading Comprehension by Generating Question and Answer Pairs}
\documentclass[runningheads,a4paper]{llncs}

\usepackage{amssymb}
\usepackage{graphicx}
\usepackage{color}
\usepackage{caption}
\usepackage{mathrsfs}
\usepackage{comment}
\usepackage[dvipsnames]{xcolor}
\usepackage{amsmath}
\usepackage{booktabs}

\usepackage[colorinlistoftodos]{todonotes}
\usepackage{algpseudocode}
\usepackage[ruled]{algorithm2e}
\usepackage{textcomp}
\usepackage[moderate]{savetrees}

\usepackage{url}
\urldef{\mailsa}\path|vishwajeet@cse.iitb.ac.in,|
\urldef{\mailsb}\path|{kireeti.boorla, yogesiitbcse}@gmail.com,|
\urldef{\mailsc}\path|ganesh@cse.iitb.ac.in,|
\urldef{\mailsd}\path|yuanfang.li@monash.edu|  

\newcommand{\keywords}[1]{\par\addvspace\baselineskip
\noindent\keywordname\enspace\ignorespaces#1}

\newcommand{\shortcite}[1]{\cite{#1}}

\newcommand{\qgf}{QG+{\color{Brown}F}}
\newcommand{\qgfne}{QG+{\color{Brown}F}+{\color{blue}NE}}
\newcommand{\qggae}{QG+{\color{blue}GAE}}
\newcommand{\qgfaes}{QG+{\color{Brown}F}+{\color{blue}AES}}
\newcommand{\qgfaeb}{QG+{\color{Brown}F}+{\color{blue}AEB}}
\newcommand{\qgfgae}{QG+{\color{Brown}F}+{\color{blue}GAE}}

\begin{document}

\mainmatter  

\title{Automating Reading Comprehension by Generating Question and Answer Pairs}

\titlerunning{Automating Reading Comprehension by Generating Question and Answer Pairs}

%
%
\author{Vishwajeet Kumar$^{\dagger,\ddagger}$ \and Kireeti Boorla$^{\dagger}$ \and Yogesh Meena$^{\dagger}$ \and\\
Ganesh Ramakrishnan$^{\dagger}$ \and Yuan-Fang Li$^{\S}$
 }
\authorrunning{Automating reading comprehension by generating question and answer pairs}

\institute{
$^\dagger$Indian Institute of Technology Bombay, India\\
$^\ddagger$IITB-Monash Research Academy, Mumbai, India\\
$^\S$Faculty of Information Technology, Monash University, Australia\\
\mailsa\\
\mailsb\\
\mailsc\\
\mailsd\\
}

%
%

\toctitle{Lecture Notes in Computer Science}
\tocauthor{Authors' Instructions}
\maketitle

\begin{abstract}
Neural network-based methods represent the state-of-the-art in question generation from text. Existing work focuses on generating only questions from text without concerning itself with answer generation. Moreover, our analysis shows that handling rare words and generating the most appropriate question given a candidate answer are still challenges facing existing approaches. 
We present a novel two-stage process to generate question-answer pairs from the text. For the first stage, we present alternatives for encoding the span of the pivotal answer in the sentence using Pointer Networks. In our second stage, we employ sequence to sequence models for question generation, enhanced with rich linguistic features. Finally, global attention and answer encoding are used for generating the question most relevant to the answer. We motivate and linguistically analyze the role of each component in our framework and consider compositions of these. This analysis is supported by extensive experimental evaluations. Using standard evaluation metrics as well as human evaluations, our experimental results validate the significant improvement in the quality of questions generated by our framework over the state-of-the-art.
The technique presented here represents another step towards more automated reading comprehension assessment. We also present a live system\footnote{Demo of the system is available at \url{https://www.cse.iitb.ac.in/~vishwajeet/autoqg.html}.} to demonstrate the effectiveness of our approach.

\keywords{Pointer Network, sequence to sequence modeling, question generation}
\end{abstract}

\section{Introduction}
Asking relevant and intelligent questions has always been an integral part of human learning, as it can help assess the user's understanding of a piece of text (an article, an essay {\em etc.}). However, forming questions manually can be sometimes  arduous. Automated question generation (QG) systems can help alleviate this problem by learning to generate questions on a large scale and in lesser time. 
Such a system  has applications in a myriad of areas such as FAQ generation, intelligent tutoring systems, and virtual assistants.

The task for a QG system is to generate meaningful, syntactically correct, semantically sound and natural questions from text. Additionally, to further automate the assessment of human users, it is highly desirable that the questions are relevant to the text and have supporting answers present in the text. 

Figure \ref{qg} below shows a sample of questions generated by our approach using a variety of configurations (vanilla sentence, feature tagged sentence and answer encoded sentence)  that will be described later in this paper. 

\begin{figure}[ht!]
\centering
\includegraphics[width=.9\linewidth]{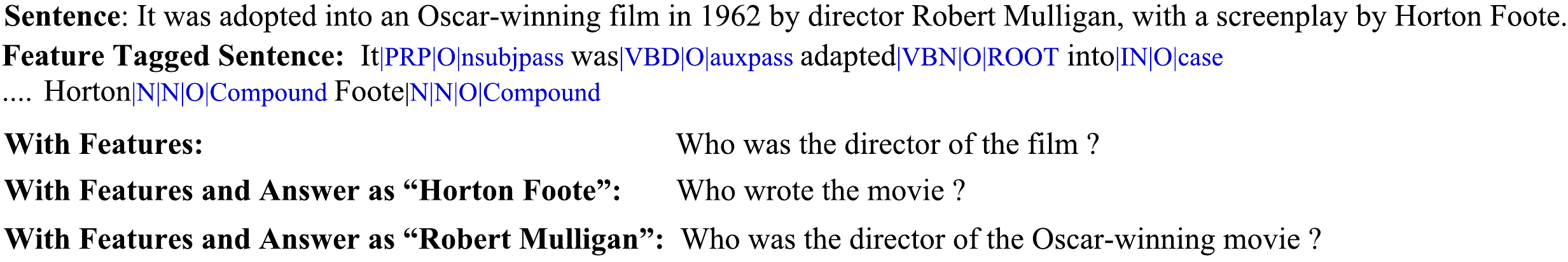}
\caption{Example: sample questions generation from text by our models.}
\label{qg}
\end{figure}

Initial attempts at automated question generation were heavily dependent on a limited, ad-hoc, hand-crafted set of rules \cite{heilman2010good,Yao2012DDqg}. These rules focus mainly on the syntactic structure of the text and are limited in their application, only to sentences of simple structures. Recently, the success of sequence to sequence learning models~\cite{sutskever2014sequence} opened up possibilities of looking beyond a fixed set of rules for the task of question generation~\cite{serban2016generating,Xinya}. When we encode ground truth answers into the sentence along with other linguistic features, we get improvement of upto 4 BLEU points along with improvement in the quality of questions generated.
A recent deep learning approach to question generation \cite{serban2016generating} investigates a simpler task of generating questions only from a triplet of subject, relation and object. In contrast, we build upon recent works that train sequence to sequence models for generating questions from natural language text.

Our work significantly improves the latest work of sequence to sequence learning based question generation using deep networks~\cite{Xinya} by making use of (i) an additional module to predict span of best answer candidate on which to generate the question
(ii) several additional rich set of linguistic features to help model generalize better (iii) suitably modified decoder to generate questions more relevant to the sentence.

The rest of the paper is organized as follows. In 
{S}ection \ref{probdef} we formally describe our question generation problem, followed by a discussion on related work in Section \ref{relWork}.
In Section~\ref{contri} we describe our approach and methodology and summarize our main contributions. In Sections~\ref{sec:answersel} and \ref{sec:quegen} we describe the two main components of our framework. Implementation details of the models are described in Section \ref{implDetails}, followed by experimental results in Section~\ref{expt} and conclusion in Section~\ref{conc}.
\section{Problem Formulation}
\label{probdef}
Given a sentence \textbf{S}, viewed as a sequence of words, our goal is to generate a question \textbf{Q}, which is syntactically and semantically correct, meaningful and natural. More formally, given a sentence \textbf{S}, our model's main objective is to learn the underlying conditional probability distribution $P(\textbf{Q}|\textbf{S};\theta)$ parameterized by $\theta$ to generate the most appropriate question that is closest to the human generated question(s).  
Our model learns $\theta$ during training using sentence/question pairs such that the probability $P(\textbf{Q}|\textbf{S};\theta$) is maximized over the given training dataset.

Let the sentence \textbf{S} be a sequence of $M$ words $(w_1, w_2, w_3, ...w_M)$, and question \textbf{Q} a sequence of $N$ words $(y_1, y_2, y_3,...y_N)$. Mathematically, the model is meant to generate 
{\textbf{Q}*} such that:
\begin{eqnarray}
\mathbf{Q^* } & = & \underset{\textbf{Q}}{\operatorname{argmax}}~P(\textbf{Q}|\textbf{S};\theta) \\
& = & \underset{y_1,..y_{n}}{\operatorname{argmax}}~\prod_{i=1}^{N}P(y_i|y_1,..y_{i-1},w_1..w_M;\theta)
\label{argmaxprobeq}
\end{eqnarray}

Equation~(\ref{argmaxprobeq}) is to be realized using a RNN-based architecture, which is described in detail in Section \ref{decoder}.

\section{Related Work}\label{relWork}
{Heilman and Smith}~\shortcite{heilman2010good} use a set of hand-crafted syntax-based rules to generate questions from simple declarative sentences. The system identifies multiple possible answer phrases from all {declarative} sentences using the constituency parse tree structure of each sentence. The system then over-generates questions and ranks them statistically by assigning scores using logistic regression.

\cite{Yao2012DDqg} use semantics of the text by converting it into the Minimal Recursion Semantics notation \cite{Copestake}. Rules specific to the summarized semantics are applied to generate questions. Most of the approaches proposed for the QGSTEC challenge \cite{mannem2010question} are also rule-based systems, some of which put to use sentence features such as part of speech (POS) tags and named entity relations (NER) tags. \cite{chali2015towards} use {ASSERT (an automatic statistical semantic role tagger that can annotate naturally occurring text with semantic arguments) for semantic role parses, generate questions based on rules and rank them based on subtopic similarity score using ESSK (Extended String Subsequence Kernel)}. \cite{ali2010automation} break sentences into fine and coarse classes and proceed to generate questions based on templates matching these classes.

All approaches mentioned so far are heavily dependent on rules whose design requires deep linguistic knowledge and yet are not exhaustive enough. Recent successes in neural machine translation \cite{sutskever2014sequence,cho2014properties} have helped address this problem by letting deep neural nets learn the implicit rules through data. This approach has inspired application of sequence to sequence learning to automated question generation. \cite{serban2016generating} propose an attention-based \cite{bahdanau2014neural,LuongPM15} approach to question generation from a pre-defined template of knowledge base triples (subject, relation, object). 
Additionally, recent studies suggest that the sharp learning capability of neural networks does not make linguistic features redundant in machine translation. \cite{features} suggest augmenting each word with its linguistic features such as POS, NER. \cite{features1} suggest a tree-based encoder to incorporate features, although for a different application.


{We build on the recent sequence to sequence learning-based method of question generation by \cite{Xinya}}, but with significant differences and improvements from all previous works in the following ways. (i) Unlike \cite{Xinya} our question generation technique is \emph{pivoted} on identification of the best candidate answer (span) around which the question should be generated. (ii) Our approach is enhanced with the use of several syntactic and linguistic features that help in learning models that generalize well. 
(iii) We propose a modified decoder to generate questions relevant to the text.

\section{Approach and Contributions}
\label{contri}

Our approach to generating question-answer pairs from text is a two-stage process: in the first stage we select the most relevant and appropriate candidate answer, i.e., the \emph{pivotal answer}, using an answer selection module, and in the second stage we encode the answer span in the sentence and use a sequence to sequence model with a rich set of linguistic features to generate questions for the pivotal answer. 

Our sentence encoder transforms the input sentence into a list of 
fixed-length continuous vector word representation, each input symbol being represented as a vector. The question decoder takes in the output from the sentence encoder  and produces one symbol at a time and stops at the EOS (end of sentence) marker. To focus on certain important words while generating questions (decoding) we use a global attention mechanism. The attention module is connected to both the sentence encoder as well as the question decoder, thus allowing the question decoder to focus on appropriate segments of the sentence while generating the next word of the question. We include linguistic features for words so that the model can learn more generalized syntactic transformations. 
We provide a detailed description of these modules in the following sections. 
Here is a summary of our 
{three} main contributions: (1) a versatile neural network-based answer selection and Question Generation (QG) approach
and an associated dataset of question/sentence pairs\footnote{Publicly available at \url{https://goo.gl/Q67cB7}.} suitable for learning answer selection, (2) incorporation of  linguistic features that help generalize the learning to syntactic and semantic transformations of the input, and (3) a modified decoder to generate the question most  relevant to the text. 

\section{Answer Selection and Encoding\label{sec:answersel}}
In applications such as reading comprehension, it is natural for a question to be generated keeping the answer in mind (hereafter referred to as the {`pivotal' answer}). Identifying the most appropriate pivotal answer will allow comprehension be tested more easily and with even higher automation. We propose a novel named entity selection model and answer selection model based on Pointer Networks
\cite{vinyals2015pointer}. These models give us the span of pivotal answer in the sentence, which we encode using the BIO notation while generating the questions.

\subsection{Named Entity Selection}\label{nesel}
In our {first} approach, we restrict our pivotal answer to be one of the named entities in the sentence, extracted using the Stanford CoreNLP toolkit.
To choose the most appropriate pivotal answer for QG from a set of 
{candidate} entities present in the sentence we propose a named entity selection model. We train a multi-layer perceptron on the sentence, named entities present in the sentence and the ground truth answer. The model learns to predict the pivotal answer given the sentence and a set of candidate entities. 
The sentence $S = (w_1, w_2, ... , w_n)$ is first encoded using a 2 layered unidirectional LSTM encoder into hidden activations $H = (h_1^s, h_2^s, ... , h_n^s)$. For a named entity $NE = (w_i, ... , w_j)$, a vector representation (\textbf{R}) is created as $<h_n^s;h_{mean}^s;h_{mean}^{ne}>$, where $h_n^s$ is the final state of the hidden activations, $h_{mean}^s$ is the mean of all the activations and $h_{mean}^{ne}$ is the mean of hidden activations $(h_i^s, ... , h_j^s)$ between the span of the named entity. This representation vector \textbf{R} is fed into a multi-layer perceptron, which predicts the probability of a named entity being a pivotal answer. Then we select the entity with the highest probability as the answer entity. More formally,
\begin{equation}
P(NE_i|S) = softmax(\textbf{R}_i.W+B)
\end{equation}
where $W$ is weight, $B$ is bias, and $P(NE_i|S)$ is the probability of named entity being the pivotal answer.

\subsection{Answer Selection using Pointer Networks}\label{answerselpointer}
We propose a novel Pointer Network ~\cite{vinyals2015pointer} based approach to find the span of pivotal answer given a sentence. Using the attention mechanism, a \emph{boundary} Pointer Network output start and end positions from the input sequence.
More formally, the problem can be formulated as follows: given a sentence \textbf{S}, we want to predict the start index $a_k^{start}$ and the end index $a_k^{end}$ of the pivotal answer. The main motivation in using a boundary pointer network is to predict the span from the input sequence as output. While we adapt the boundary pointer network to predict the start and end index positions of the pivotal answer in the sentence, we also present results using a sequence pointer network instead. 
\noindent \textbf{Answer sequence pointer network} produces a sequence of pointers as output. Each pointer in the sequence is word index of some token in the input. It only ensures that output is contained in the sentence but isn't necessarily a substring. Let the encoder's hidden states be $H = (h_1,h_2,\ldots,h_n)$ for a sentence the probability of generating output sequence $O$ = $(o_1,o_2,\ldots,o_m)$ is defined as, 
\begin{equation}
P(O|S) = \prod P(o_i|o_1,o_2,o_3,\ldots,o_{i-1},H)
\end{equation}
We model the probability distribution as:
\begin{equation}
u^i = v^T tanh(W^e\hat{H}+W^dD_i)
\end{equation}
\begin{equation}
P(o_i|o_1,o_2,\ldots.,o_{i-1},H) = softmax(u^i)
\end{equation}

Here, $W^e\in R^{d \times 2d}$, $W^D\in R^{d \times d}$, $v\in R^d$ are the model parameters to be learned. $\hat{H}$ is ${<}H;0{>}$, where a 0 vector is concatenated with LSTM encoder hidden states to produce an end pointer token. $D_i$ is produced by taking the last state of the LSTM decoder with inputs ${<}softmax(u^i)\hat{H};D_{i-1}{>}$. $D_0$ is a zero vector denoting the start state of the decoder. 

\begin{figure*}
\centering
\includegraphics[width=.8\linewidth]{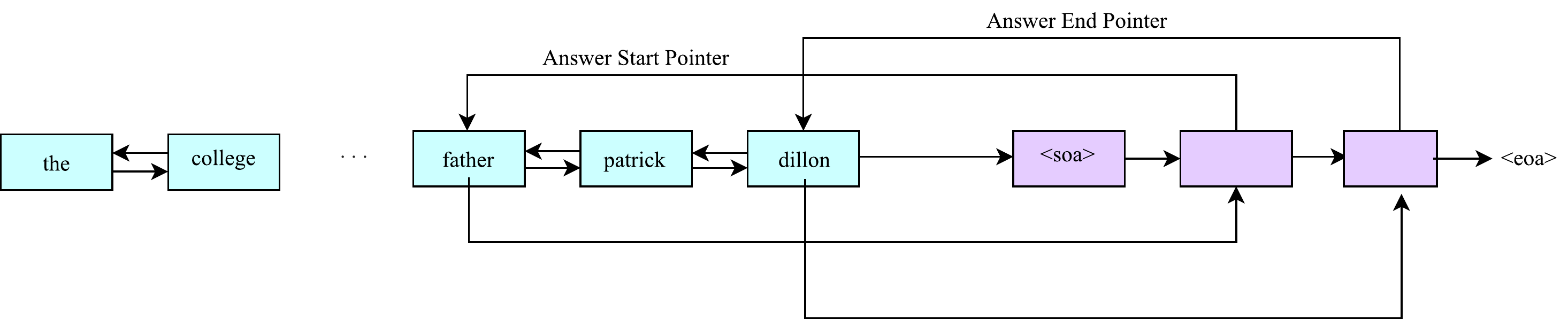}
\caption{Answer selection using boundary pointer network.}\label{fig:pn}
\end{figure*}

\noindent \textbf{Answer boundary pointer network} produces two tokens corresponding to the start and end index of the answer span. The probability distribution model remains exactly the same as answer sequence pointer network. The boundary pointer network is depicted in Figure~\ref{fig:pn}.

We take sentence S = $(w_1,w_2,\ldots,w_M)$ and generate the hidden activations H by using embedding lookup and an LSTM encoder. As the pointers are not conditioned over a second sentence, the decoder is fed with just a start state.

\textbf{Example:}
For the Sentence: ``\texttt{other past residents include composer journalist and newspaper editor william henry wills , ron goodwin , and journalist angela rippon and comedian dawn french}'', the answer pointers produced are:

\noindent\textbf{Pointer(s) by answer sequence:} [6,11,20] $\rightarrow$ journalist henry rippon 

\noindent\textbf{Pointer(s) by answer boundary:} [10,12] $\rightarrow$ william henry wills

\section{Question Generation\label{sec:quegen}}
After encoding the pivotal answer (prediction of the answer selection module) in a sentence, we train a sequence to sequence model augmented with a rich set of linguistic features to generate the question. In sections below we describe our linguistic features as well as our sequence to sequence model.

\subsection{Sequence to Sequence Model}\label{seqtoseq}
Sequence to sequence models \cite{sutskever2014sequence} learn to map input sequence (sentence) to an intermediate fixed length vector representation using an encoder RNN along with the mapping for translating this vector representation to the output sequence (question) using another decoder RNN. 
Encoder of the sequence to sequence model first conceptualizes the sentence as a single fixed length vector before passing this along to the decoder which uses this vector and attention weights to generate the output.\\
\\

\noindent \textbf{Sentence Encoder:}
The sentence encoder is realized using a bi-directional LSTM. In the forward pass, the given sentence along with the linguistic features is fed through a recurrent activation function recursively till the whole sentence is processed. Using one LSTM as encoder will capture only the left side sentence dependencies of the current word being fed. To alleviate this and thus to also capture the right side dependencies of the sentence for the current word while predicting in the decoder stage, another LSTM is fed with the sentence in the reverse order. The combination of both is used as the encoding of the given sentence.  \begin{equation} 
\overrightarrow{\hat{h}_t}=f(\overrightarrow{W}w_t + \overrightarrow{V}\overrightarrow{\hat{h}_{t-1}} +\overrightarrow{b}) 
\end{equation}
\begin{equation}
\overleftarrow{\hat{h}_t}=f(\overleftarrow{W}w_t + \overleftarrow{V}\overleftarrow{\hat{h}_{t+1}} +\overleftarrow{b}) 
\end{equation}
\begin{equation}
\hat{h_t}=g(Uh_t + c) = g(U[\overrightarrow{\hat{h}_t},\overleftarrow{\hat{h}_t}] + c)
\end{equation}

The hidden state $\hat{h_t}$ of the sentence encoder is used as the intermediate representation of the source sentence at time step $t$ whereas $W, V, U \in R^{n\times m}$ are weights, where \textit{m} is the word embedding dimensionality, \textit{n} is the number of hidden units, and $w_t \in R^{p\times q \times r} $ is the weight vector corresponding to feature encoded input at time step $t$.

\noindent \textbf{Attention Mechanism:}
\label{atten}
  In the commonly used sequence to sequence model (\cite{sutskever2014sequence}), the decoder is directly initialized with  intermediate source representation ($\hat{h_t}$). Whereas the attention mechanism proposed in \cite{LuongPM15} suggests using a subset of source hidden states, giving more emphasis to a, possibly, more relevant part of the context in the source sentence while predicting a new word in the target sequence. In our method we specifically use the global attention mechanism. 
In this mechanism a context vector $c_t$ is generated by capturing relevant source side information for predicting the current target word $y_t$ in the decoding phase at time $t$. Relevance between the current decoder hidden state $h_t$ and each of the source hidden states ($\hat{h_1},\hat{h_2}...\hat{h_{N}}$) is realized through a dot similarity metric: $score(h_t,\hat{h_i}) = h_t^{T}\cdot\hat{h_i}$.

A softmax layer (\ref{attention_weights}) is applied over these scores to get the variable length alignment vector $\alpha_t$ which in turn is used to compute the weighted sum  over all the source hidden states ($\hat{h_1},\hat{h_2}, \ldots, \hat{h_N}$) to generate the context vector $c_t$ (\ref{contextvector}) at time $t$. 

\begin{align}
\alpha_t(i) &= align(h_t,\hat{h_i}) 
			=\frac{\exp(score(h_t,\hat{h_i}) }{\sum\limits_{i'} \exp(score(h_t,\hat{h_{i'}}))}\label{attention_weights}\\
%
c_t &= \sum\limits_{i} \alpha_{ti} \hat{h_i}\label{contextvector}
\end{align}

\subsubsection{Question Decoder:}
\label{decoder}

Question decoder is a two layer LSTM network. It takes output of sentence encoder and decodes it to generate question.
The question decoder is designed to maximize our objective in equation \ref{argmaxprobeq}. More formally decoder computes probability $P(Q|S;\theta)$ as:
\begin{equation}
P(Q|S;\theta)=softmax(W_s(tanh(W_r[h_t,c_t]+b)))
\end{equation}
where $W_s$ and $W_r$ are weight vectors and \textit{tanh} is the activation function.
The hidden state of the decoder along with the context vector $c_t$ is used to predict the target word $y_t$.
It is a known fact that decoder may output words which are not even present in the source sentence as it learns a probability distribution over the words in the vocabulary. To generate questions relevant to the text we suitably modified decoder and integrated an attention mechanism (described in Section \ref{atten}) with the decoder to attend to words in source sentence while generating questions. This modification to the decoder increases the relevance of question generated for a particular sentence. 
\subsection{Linguistic Features \label{sec:linguisticfeat}}
We propose using a set of linguistic features so that the model can learn better generalized transformation rules, rather than learning a transformation rule per sentence. We describe our features below:

\noindent \textbf{POS Tag:} Parts of speech tag of the word. Words having same POS tag have similar grammatical properties and demonstrate similar syntactic behavior. We use the Stanford ConeNLP -pos annotator to get POS Tag of words in the sentence.

\noindent \textbf{Named Entity Tag:} Name entity tag represent coarse grained category of a word for example PERSON, PLACE, ORGANIZATION, DATE, etc. In order to help the model identify named entities present in the sentence, named entity tag of each word is provided as a feature. This ensures that the model learns to pose a question about the entities present in the sentence. We use the Stanford CoreNLP -ner annotator to assign named entity tag to each word. 

\noindent \textbf{Dependency Label:} Dependency label of a word is the edge label connecting each word with the parent in the dependency parse tree. Root node of the tree is assigned label `ROOT'. Dependency label help models to learn inter-word relations. It helps in understanding the semantic structure of the sentence while generating question. Dependency structure also helps in learning syntactic transformations between sentence and question pair. Verbs and adverbs present in the sentence signify the type of the question (which, who .. etc.) that would be posed for the subject it refers to.
We use dependency parse trees generated using the Stanford CoreNLP parser to obtain the dependency labels. 

Linguistic features are added by the conventional feature concatenation of tokens using the delimiter `$|$'. We create separate vocabularies for words (encoded using glove's pre-trained word embedding) and features (using one-hot encoding) respectively.


\section{Implementation Details\label{implDetails}}
We implement our answer selection and question generation models in Torch\footnote{\url{http://torch.ch/}}. The sentence encoder of \textbf{QG} is a 3 layer bi-directional LSTM stack and the question decoder is a 3 layer LSTM stack. Each LSTM has a hidden unit of size 600 units. we use pre-trained glove embeddings\footnote{\url{http://nlp.stanford.edu/data/glove.840B.300d.zip}}~\cite{pennington2014glove} of 300 dimensions for both the encoder and the decoder. All model parameters are optimized using Adam optimizer with a learning rate of 1.0 and we decay the learning rate by 0.5 after 10th epoch of training. The dropout probability is set to 0.3. We train our model in each experiment for 30 epochs, we select the model with the lowest perplexity on validation set. 
\par 
The linguistic features for each word such as POS, named entity tag {\em etc.}, are incorporated along with word embeddings through concatenation. 


\section{{Experiments and Results}\label{expt}}

We evaluate performance of our models on the SQUAD \cite{rajpurkar2016squad} dataset (denoted \textbf{$\mathcal{S}$}). We use the same split as that of \cite{Xinya}, where a random subset of 70,484 instances from \textbf{$\mathcal{S}\ $} are used for training (${\mathcal S}^{tr}$), 10,570 instances for validation (${\mathcal S}^{val}$), and 11,877 instances for testing (${\mathcal S}^{te}$). 

We performed both human-based evaluation as well as automatic evaluation to assess the quality of the questions generated. 
For automatic evaluation, we report results using a metric widely used to evaluate machine translation systems, called BLEU~\cite{papineni2002bleu}.

We first list the different systems (models) that we evaluate and compare in our experiments. A note about abbreviations: Whereas components in {\color{blue} blue} are different alternatives for encoding the pivotal answer, the  {\color{Brown} brown} color coded component represents the set of linguistic features that can be optionally added to any model.

\noindent \textbf{Baseline System (QG):} 
Our baseline system is a sequence-to-sequence LSTM model (see Section~\ref{sec:quegen}) trained only on raw sentence-question pairs without using features or answer encoding. This model is the same as \cite{Xinya}.

\noindent \textbf{System with feature tagged input (\qgf{}):}
We encoded linguistic features (see Section~\ref{sec:linguisticfeat}) for each sentence-question pair to augment the basic \textbf{QG} model. This was achieved by appending features to each word using the ``$|$'' delimiter. 
This model helps us analyze the isolated effect of incorporating syntactic and semantic properties of the sentence (and words in the sentence) on the outcome of question generation. 

\noindent \textbf{Features + NE encoding (\qgfne{}):}
We also augmented the feature-enriched sequence-to-sequence \qgf{} model by encoding each named entity predicted by the named entity selection module (see section \ref{nesel}) as a pivotal answer. This model helps us analyze the effect of (indiscriminate) use of named entity as potential (pivotal) answer, when used in conjunction with features.

\noindent \textbf{Ground truth answer encoding (\qggae{}):} In this setting we use the  encoding of ground truth answers from sentences to augment the training of the basic QG model (see Section~\ref{sec:answersel}). For encoding answers into the sentence we employ the BIO notation. We append ``\textbf{B}'' as a feature using the delimiter ``$|$'' to the first word of the answer and ``\textbf{I}'' as a feature for the rest of the answer words. 
We used this model to analyze the effect of answer encoding on question generation, independent of features and named entity alignment. 

We would like to point  out that any direct comparison  of a generated question with the question in the ground truth using any machine translation-like metric (such as the BLEU metric discussed in Section~\ref{sec:results}) makes sense only when both the questions are associated with the same pivotal answer. This specific experimental setup and the ones that follow are therefore more amenable for evaluation using standard metrics used in machine translation.

\noindent \textbf{Features + sequence pointer network predicted answer encoding (\qgfaes{}):}
In  this setting, we encoded the pivotal answer in the sentence as predicted by the sequence pointer network (see Section~\ref{answerselpointer}) to augment the  linguistic feature based \textbf{\qgf{}} model. In this and in the following setting, we
expect the prediction of the pivotal answer in the sentence to closely approximate the ground truth answer. 

\noindent \textbf{Features + boundary pointer network predicted answer encoding (\qgfaeb{}):}
In  this setting, we encoded the pivotal answer in the sentence as predicted by the boundary pointer network (see Section~\ref{answerselpointer}) to augment the  linguistic feature based \textbf{\qgf{}} model.

\noindent \textbf{Features + ground truth answer encoding (\qgfgae{}):}
In this experimental setup, building upon the previous model (\qgf{}), we encoded ground truth answers to augment the QG model.   

\subsection{Results and Analysis \label{sec:results}}
We compare the performance of the $7$ systems QG,  \qgf{}, \qgfne{},  \qggae{},  \qgfaes{}, \qgfaeb{} and \qgfgae{} described in the previous sections on (the train-val-test splits of) ${\mathcal S}$ and report results using both human and automated evaluation metrics. We first describe experimental results using human evaluation followed by evaluation on other metrics.

\noindent  \textbf{Human Evaluation:}
We randomly selected 100 sentences from the test set (${\mathcal S}^{te}$) and generated one question using each of the $7$ systems for each of these 100 sentences and asked three human experts for feedback on the quality of questions generated. Our human evaluators are professional English language experts. They were asked to provide feedback about a randomly sampled sentence along with the corresponding questions from each competing system, presented in an anonymised random order. This was to avoid creating any bias in the evaluator towards any particular system. They were not at all primed about the different models and the hypothesis.

We asked the following binary (yes/no) questions to each of the experts: a) \textsf{is this question syntactically correct?}, b) \textsf{is this question semantically correct?}, and c) \textsf{is this question relevant to this sentence?}. Responses from all three experts were collected and averaged. For example, suppose  the cumulative scores of the 100 binary judgements for syntactic correctness  by the 3 evaluators were $(80, 79, 73)$. Then the average response would be 77.33. 
In Table~\ref{HESQUAD} we present these results on the test set ${\mathcal S}^{te}$.

\begin{table}[htp]
\centering
\setlength{\tabcolsep}{4pt}
\renewcommand{\arraystretch}{0.8}
\begin{tabular}{lrrr}
\toprule
\textbf{System} & Syntactically correct (\%) & Semantically correct (\%) & Relevant (\%) \\
\midrule
QG \cite{Xinya} & 51.6 & 48 &52.3 \\
\midrule
\qgf{} & 59.6 & 57 & 64.6 \\
\midrule
\qgfne{} & 57 &  52.6 & 67\\
\midrule
\qggae{} & 44 & 35.3 & 50.6\\
\midrule
\qgfaes{} & 51 & 47.3 & 55.3\\
\midrule
\textbf{\qgfaeb{}} & 61 & 60.6 & \textbf{71.3}\\
\midrule
\textbf{\qgfgae{}} & \textbf{63} & \textbf{61} & 67 \\
\bottomrule

\end{tabular}
\caption{Human evaluation results on  ${\mathcal S}^{te}$. Parameters are, \textbf{p1}: percentage of syntactically correct questions, \textbf{p2}: percentage of semantically correct questions, \textbf{p3}: percentage of relevant questions.}
\label{HESQUAD}
\end{table}

\begin{table}[htp]
\centering
\setlength{\tabcolsep}{3pt}
\renewcommand{\arraystretch}{0.8}
\begin{tabular}{l*{6}{r}}
\toprule
\textbf{Model} & \textbf{BLEU-1} & \textbf{BLEU-2} &  \textbf{BLEU-3} & \textbf{BLEU-4} & \textbf{METEOR} & \textbf{ROUGE-L} \\
\midrule
QG \cite{Xinya} & 39.97 &22.39 & 14.39 & 9.64 & 14.34 & 37.04 \\ 
\midrule
\qgf{} & 41.89 & 24.37 &15.92  & 10.74 & 15.854 & 37.762 \\ 
\midrule
\qgfne{} &41.54 & 23.77 & 15.32& 10.24 &15.906 & 36.465 \\ 
\midrule
\qggae{} &43.35 & 24.06 & 14.85 & 9.40 & 15.65 & 37.84 \\ 
\midrule
\textbf{\qgfaes{}} &\textbf{43.54} & \textbf{25.69} & \textbf{17.07} & \textbf{11.83} & \textbf{16.71} & \textbf{38.22} \\ 
\midrule
\textbf{\qgfaeb{}} &\textbf{42.98} & \textbf{25.65} & \textbf{17.19} & \textbf{12.07} & \textbf{16.72} & \textbf{38.50} \\ 
\midrule 
\textbf{\qgfgae{}} &\textbf{46.32} & \textbf{28.81} & \textbf{19.67} & \textbf{13.85} & \textbf{18.51} & \textbf{41.75}\\ 
\bottomrule
\end{tabular}
\caption{Automatic evaluation results on ${\mathcal S}^{te}$. BLEU, METEOR and ROUGE-L scores vary between 0 and 100, with the upper bound of 100 attainable on the ground truth. QG\cite{Xinya}:Result obtained using latest version of Torch.}
\label{results}
\end{table}


\noindent \textbf{Evaluation on other metrics:}
We also evaluated our system on other standard metrics to enable comparison with other systems. However, as explained earlier, the standard metrics used in machine translation such as BLEU \cite{papineni2002bleu}, METEOR \cite{denkowski:lavie:meteor-wmt:2014}, and ROUGE-L \cite{lin:2004:ACLsummarization}, might not be appropriate measures to evaluate the task of question generation. To appreciate this, consider the candidate question ``who was the widow of mcdonald 's owner ?'' against the ground truth ``to whom was john b. kroc married ?'' for the sentence ``\texttt{it was founded in 1986 through the donations of joan b. kroc , the widow of mcdonald 's owner ray kroc.}''. \allowbreak It is easy to see that the candidate is a valid question and makes perfect sense. However its BLEU-4 score is almost zero. 
Thus, it may be the case that the human generated question against which we evaluate the system generated questions may be completely different in structure and semantics, but still be perfectly valid, as seen previously. While we find human evaluation to be  more appropriate, for the sake of completeness, we also report the BLEU, METEOR and ROUGE-L scores in each setting.In Table \ref{results}, we observe that our models, \qgfaeb{}, \qgfaes{} and \qgfgae{} outperform the state-of-the art question generation system QG~\cite{Xinya} significantly on all standard metrics. 

Our model \qgfgae{}, which encodes ground truth answers and uses a rich set of linguistic features,  performs the best as per every metric. And in Table \ref{HESQUAD}, we observe that adding the rich set of linguistic features to the baseline model (QG) further improves performance. Specifically, addition of features increases syntactic correctness of questions by 2\%, semantic correctness by 9\% and relevance of questions with respect to sentence by 12.3\% in comparison with the baseline model QG \cite{Xinya}.
\begin{figure}[htp]
\centering
\includegraphics[width=.9\linewidth]{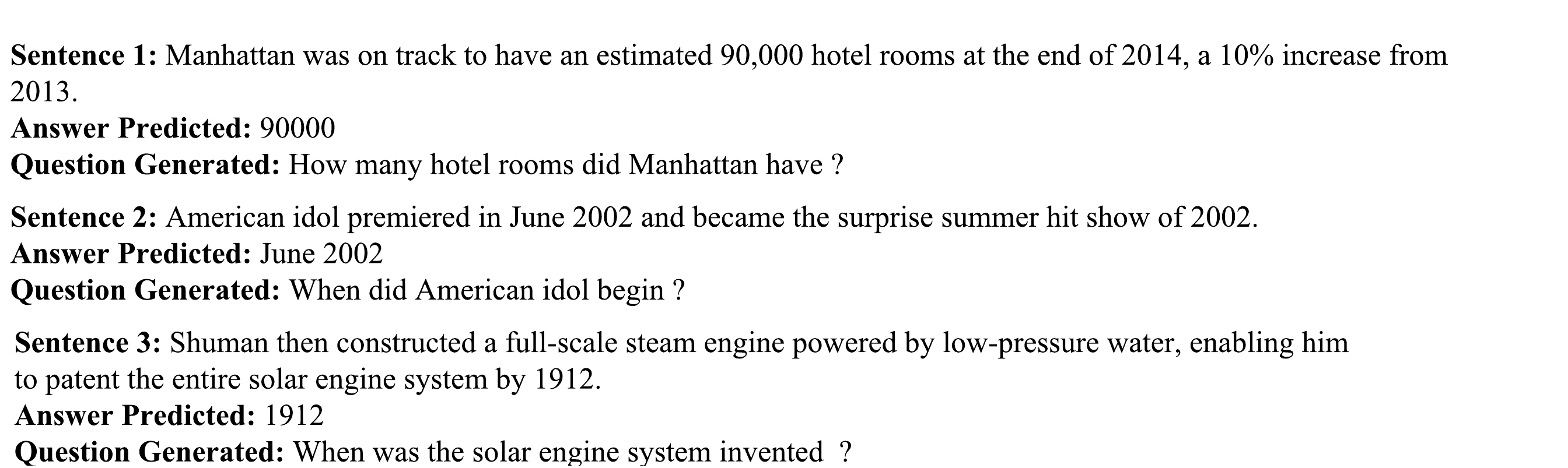}
\caption{Sample output: the pivotal answer predicted and the question generated about the answer using model \qgfaeb{} - that is comparable to the best performing system that also used ground truth answers.}
\label{sample_output}
\end{figure}

In Figure \ref{sample_output} we present some sample answers predicted and corresponding questions generated by our model \qgfaeb{}. Though not better, the performance of models \qgfaes{} and \qgfaeb{} is comparable to the best model (that is \qgfgae{}, which additionally uses ground truth answers).  This is because the ground truth answer might not be the best and most relevant pivotal answer for question generation, particularly since each question in the SQUAD dataset was generated by looking at an entire paragraph and not any single sentence. Consider the sentence ``\textit{manhattan was on track to have an estimated 90,000 hotel rooms at the end of 2014 , a 10 \% increase from 2013 .}''. On encoding the ground truth answer, ``\textit{90,000}'',  the question generated using model \qggae{} is ``\textit{what was manhattan estimated hotel rooms in 2014 ?}'' and 
and additionally, with linguistic features (\qgfgae{}),  we get ``\textit{how many hotel rooms did manhattan have at the end of 2014 ?}''. This is indicative of how a rich set of linguistic features help in shaping the correct question type as well generating syntactically and semantically correct question. Further when we do not encode any answer (either pivotal answer predicted by sequence/boundary pointer network or ground truth answer) and just augment the linguistic features (\qgf{}) the question generated is ``\textit{what was manhattan 's hotel increase in 2013 ?}'', which  is clearly a poor quality question. 
Thus, both answer encoding and augmenting rich set of linguistic features are important for generating high quality (syntactically correct, semantically correct and relevant) questions. 
When we select pivotal answer from amongst the set of named entities present in the sentence ({\em i.e.}, model \qgfne{}),  the question generated on encoding the named entity ``\textit{manhattan}'' 
is ``\textit{what was the 10 of hotel 's city rooms ?}'', which is clearly a poor quality question. The poor performance of \qgfne{} can be attributed to the fact that only 50\% of the answers in SQUAD dataset are named entities.

\section{Conclusion\label{conc}}
We introduce a novel two-stage process to generate question-answer pairs from text. We combine and enhance a number of techniques including sequence to sequence models, Pointer Networks, named entity alignment, as well as rich linguistic features to identify potential answers from text, handle rare words, and generate questions most relevant to the answer. To the best of our knowledge this is the first attempt in generating question-answer pairs.  
Our comprehensive evaluation shows that our approach significantly outperforms current state-of-the-art question generation techniques on both human evaluation and evaluation on common metrics such as BLEU, METEOR, and ROUGE-L.


 {\small

\begin{thebibliography}{10}
\providecommand{\url}[1]{\texttt{#1}}
\providecommand{\urlprefix}{URL }

\bibitem{ali2010automation}
Ali, H., Chali, Y., Hasan, S.A.: Automation of question generation from
  sentences. In: 3rd Workshop on Question Generation. pp. 58--67 (2010)

\bibitem{bahdanau2014neural}
Bahdanau, D., Cho, K., Bengio, Y.: Neural machine translation by jointly
  learning to align and translate. arXiv preprint arXiv:1409.0473  (2014)

\bibitem{chali2015towards}
Chali, Y., Hasan, S.A.: Towards topic-to-question generation. Computational
  Linguistics  41(1),  1--20 (2015)

\bibitem{cho2014properties}
Cho, K., Van~Merri{\"e}nboer, B., Bahdanau, D., Bengio, Y.: On the properties
  of neural machine translation: Encoder-decoder approaches. arXiv preprint
  arXiv:1409.1259  (2014)

\bibitem{Copestake}
Copestake, A., Flickinger, D., Sag, I.A., Pollard, C.: Minimal recursion
  semantics: an introduction (1999),
  \url{http://www-csli.stanford.edu/~sag/sag.html}, draft

\bibitem{denkowski:lavie:meteor-wmt:2014}
Denkowski, M., Lavie, A.: Meteor universal: Language specific translation
  evaluation for any target language. In: EACL 2014 Workshop
  on Statistical Machine Translation (2014)

\bibitem{Xinya}
Du, X., Shao, J., Cardie, C.: Learning to ask: Neural question generation for
  reading comprehension. In: 55th Annual Meeting of the ACL. vol.~1, pp. 1342--1352 (2017)

\bibitem{features1}
Eriguchi, A., Hashimoto, K., Tsuruoka, Y.: Tree-to-sequence attentional neural
  machine translation. CoRR  abs/1603.06075 (2016),
  \url{http://arxiv.org/abs/1603.06075}

\bibitem{heilman2010good}
Heilman, M., Smith, N.A.: Good question! statistical ranking for question
  generation. In: HLT-NAACL 2010.
  pp. 609--617. Association for Computational Linguistics (2010)

\bibitem{lin:2004:ACLsummarization}
Lin, C.Y.: Rouge: A package for automatic evaluation of summaries. In:
  ACL-04 Workshop on Text Summarization Branches Out. pp. 74--81. ACL (2004)

\bibitem{LuongPM15}
Luong, M., Pham, H., Manning, C.D.: Effective approaches to attention-based
  neural machine translation. CoRR  abs/1508.04025 (2015),
  \url{http://arxiv.org/abs/1508.04025}

\bibitem{mannem2010question}
Mannem, P., Prasad, R., Joshi, A.: Question generation from paragraphs at
  upenn: Qgstec system description. In: 3rd Workshop on Question Generation. pp. 84--91 (2010)

\bibitem{papineni2002bleu}
Papineni, K., Roukos, S., Ward, T., Zhu, W.J.: Bleu: a method for automatic
  evaluation of machine translation. In: 40th annual meeting
  of the ACL. pp. 311--318. ACL (2002)

\bibitem{pennington2014glove}
Pennington, J., Socher, R., Manning, C.D.: Glove: Global vectors for word
  representation. In: Empirical Methods in Natural Language Processing (EMNLP).
  pp. 1532--1543 (2014)

\bibitem{rajpurkar2016squad}
Rajpurkar, P., Zhang, J., Lopyrev, K., Liang, P.: Squad: 100,000+ questions for
  machine comprehension of text. arXiv preprint arXiv:1606.05250  (2016)

\bibitem{features}
Sennrich, R., Haddow, B.: Linguistic input features improve neural machine
  translation. CoRR  abs/1606.02892 (2016),
  \url{http://arxiv.org/abs/1606.02892}

\bibitem{serban2016generating}
Serban, I.V. et al: Generating factoid questions with recurrent neural
  networks: The 30m factoid question-answer corpus. arXiv preprint
  arXiv:1603.06807  (2016)

\bibitem{sutskever2014sequence}
Sutskever, I., Vinyals, O., Le, Q.V.: Sequence to sequence learning with neural
  networks. In: Advances in neural information processing systems. pp.
  3104--3112 (2014)

\bibitem{vinyals2015pointer}
Vinyals, O., Fortunato, M., Jaitly, N.: Pointer networks. In: NIPS. pp.
  2692--2700 (2015)

\bibitem{Yao2012DDqg}
Yao, X., Bouma, G., Zhang, Y.: Semantics-based question generation and
  implementation. Dialogue and Discourse, Special Issue on Question Generation
  3(2),  11--42 (2012)

\end{thebibliography}
 \bibliographystyle{splncs03}

 }

\end{document}